\def\year{2020}\relax
\documentclass[letterpaper]{article} % DO NOT CHANGE THIS
\usepackage{aaai20}  % DO NOT CHANGE THIS
\usepackage{times}  % DO NOT CHANGE THIS
\usepackage{helvet} % DO NOT CHANGE THIS
\usepackage{courier}  % DO NOT CHANGE THIS
\usepackage[hyphens]{url}  % DO NOT CHANGE THIS
\usepackage{graphicx} % DO NOT CHANGE THIS
\usepackage{graphicx}  % DO NOT CHANGE THIS
\usepackage{booktabs}
\usepackage{amsfonts}
\usepackage{subfig}
\usepackage{tikz}
\usepackage{pgfplots}
\nocopyright
\title{SK-Net: Deep Learning on Point Cloud via End-to-end Discovery of Spatial Keypoints}
\author{Weikun Wu,\textsuperscript{\rm 1}\footnotemark[1] Yan Zhang,\textsuperscript{\rm 12}\thanks{Contributed equally.} David Wang,\textsuperscript{\rm 3} Yunqi Lei\textsuperscript{\rm 1}\thanks{Corresponding author.}\\ 
\textsuperscript{\rm 1}Computer Science Department, Xiamen University, China\\
\textsuperscript{\rm 2}School of Mathematics Science, Guizhou Normal University, China\\
\textsuperscript{\rm 3}Department of Electrical and Computer Engineering, The Ohio State University, USA\\
wuweikun@stu.xmu.edu.cn, zy@gznu.edu.cn,  Wang.8552@buckeyemail.osu.edu, yqlei@xmu.edu.cn\\
}

\begin{document}

\maketitle

\begin{abstract}
Since the PointNet was proposed, deep learning on point cloud has been the concentration of intense 3D research. However, existing point-based methods usually are not adequate to extract the local features and the spatial pattern of a point cloud for further shape understanding. This paper presents an end-to-end framework, \emph{SK-Net}, to jointly optimize the inference of spatial keypoint with the learning of feature representation of a point cloud for a specific point cloud task. One key process of SK-Net is the generation of spatial keypoints (Skeypoints). It is jointly conducted by two proposed regulating losses and a task objective function without knowledge of Skeypoint location annotations and proposals. Specifically, our Skeypoints are not sensitive to the location consistency but are acutely aware of shape. Another key process of SK-Net is the extraction of the local structure of Skeypoints (detail feature) and the local spatial pattern of normalized Skeypoints (pattern feature). This process generates a comprehensive representation, pattern-detail (PD) feature, which comprises the local detail information of a point cloud and reveals its spatial pattern through the part district reconstruction on normalized Skeypoints. Consequently, our network is prompted to effectively understand the correlation between different regions of a point cloud and integrate contextual information of the point cloud. In point cloud tasks, such as classification and segmentation, our proposed method performs better than or comparable with the state-of-the-art approaches. We also present an ablation study to demonstrate the advantages of SK-Net.   
\end{abstract}

\section{Introduction}
Rapid development in stereo sensing technology has made 3D data ubiquitous. The naive structure of most 3D data obtained by the stereo sensor is a point cloud. It makes those methods directly consuming points the most straightforward approach to handle the 3D tasks. Most of the existing point-based methods work on the following aspects: (1) Utilizing the multi-layer perceptron (MLP) to extract the point features and using the symmetry function to guarantee the permutation invariance. (2) Capturing the local structure through explicit or other local feature extraction approaches, and/or modeling the spatial pattern of a point cloud by discovering a set of keypoint-like points. (3) Aggregating features to deal with a specific point cloud task.

Despite being a pioneer in this area, PointNet\cite{pointnet} only extracts the point features to acquire the global representation of a point cloud. For boosting point-based performance, the study advancing on above (2) has recently fostered the proposal of many techniques.  One of these proposals is downsampling a point cloud to choose a set of spatial locations of a point cloud as local feature extraction regions\cite{pointnet++,jiang2018pointsift,liu2018point2sequence}. However, in this method, those locations are mostly obtained by artificial definition, e.g., FPS algorithm\cite{eldar1997farthest}, and do not reckon with the spatial pattern and the correlation between different regions of a point cloud. Similarly, \cite{xie2018attentional,hua2018pointwise} achieve pointwise local feature extraction by corresponding each point to a region with respect to modeling the local structure of a point cloud. A special way is that \cite{le2018pointgrid,pointcnn} combine the regular grids of 3D space with points, leveraging spatially-local correlation of regular grids to represent local information. However, performing pointwise local feature extraction and incorporating regular grids often require high computational costs, which is infeasible with large point injections and higher grid resolutions.
Furthermore, SO-Net\cite{so_net} introduces the Self-Organizing Map (SOM)\cite{som} to produce a set of keypoint-like points for modeling the spatial pattern of a point cloud. Even though SO-Net takes the regional correlation of a point cloud into account, it trains SOM separately. This leaves the spatial modeling of SOM and a specific point cloud task disconnected.

To address these issues, this paper explores a method that can benefit from an end-to-end framework while the inference of Skeypoints is jointly optimized with the learning of the local details and the spatial pattern of a point cloud. By this means, those end-to-end learned Skeypoints are in strong connection with the local feature extraction and the spatial modeling of a point cloud. It is worth mentioning that related approaches of end-to-end learning of 3D keypoints usually require the keypoint location proposals\cite{georgakis2018end,endtoend2} and expect the location consistency of 3D keypoints. Besides, most of the existing methods handle the tasks of 3D matching or 3D pose estimation. But those methods are not extended to the point cloud recognition tasks\cite{georgakis2018end,keypointnet,zhou2018unsupervised,End-to-end learning of 3D keypoints}. Our work focuses on point cloud recognition tasks. Before producing the PD feature, all local features perform max-pooling to ensure that the local spatial pattern is of permutation invariance and fine shape-description. Thus the generated Skeypoints are not sensitive to location consistency. 
With further training under the adjustment of two regulating losses, these Skeypoints gradually spread to the entire space of a point cloud and distribute around the discriminative regions of the point cloud. 
The generation of Skeypoints and the preliminary spatial exploration of the point cloud is shown in Figure \ref{train}. The \textbf{key contributions} of this paper are as follows:
\begin{figure}[t]
	\centering
	\includegraphics[width=0.85\columnwidth]{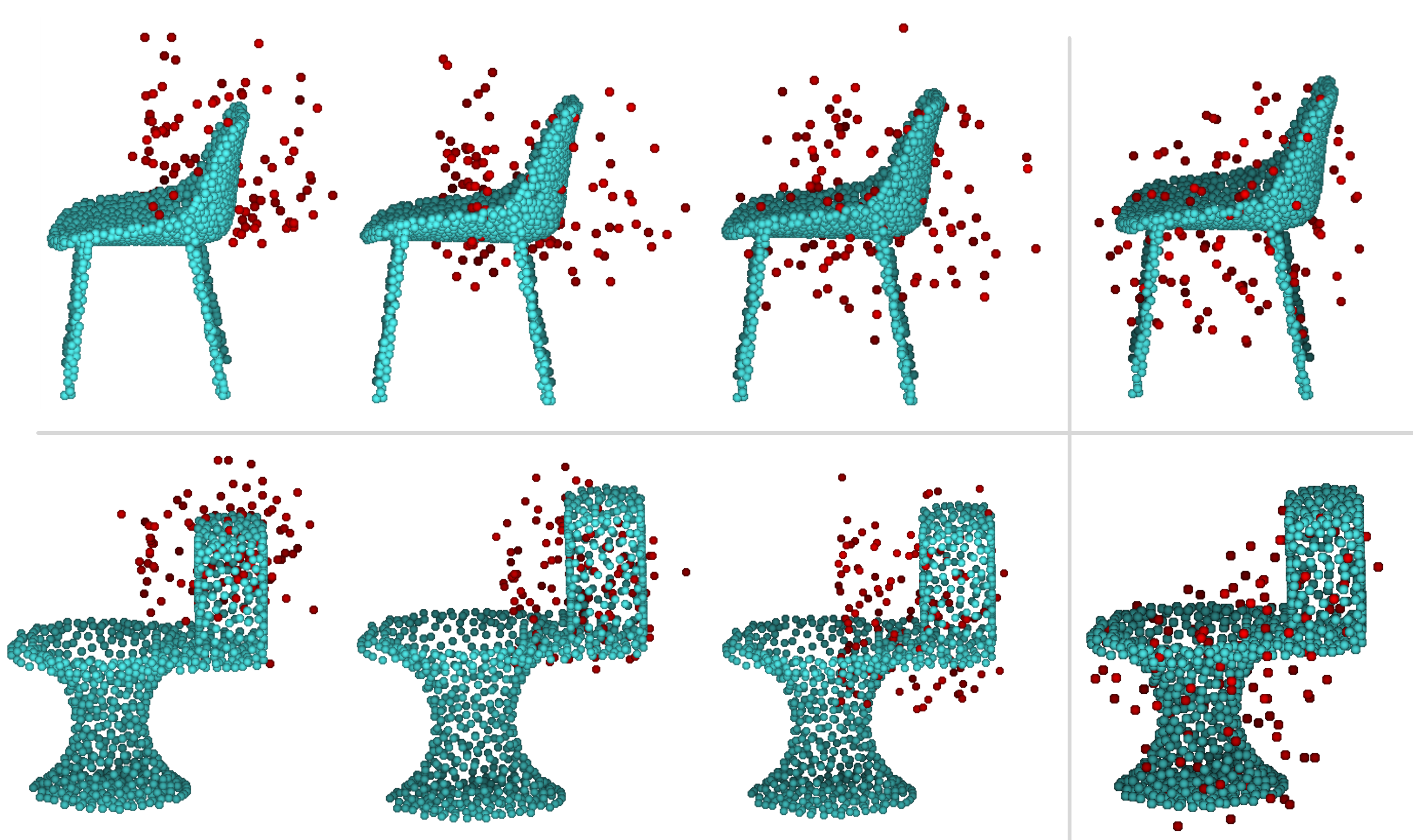} 
	\caption{The generation of Skeypoints and the preliminary spatial exploration of the point cloud. From left to right, the figures show the distributions of generated Skeypoints (red balls) with training epochs increasing; the rightmost figures show the resulting Skeypoints.}
	\label{train}
\end{figure}
\begin{itemize}
	\item  We propose a novel network named \emph{SK-Net}. It is an end-to-end framework that jointly optimizes the inference of spatial keypoints (\emph{Skeypoints}) and the learning of feature representation in the process of solving a specific point cloud task, e.g., classification and segmentation. The generated Skeypoints do not require location consistency but are acutely aware of shape. It benefits the local detail extraction and the spatial modeling of a point cloud.
	\item We design a pattern and detail extraction module (\emph{PDE module}) where the key detail features and pattern features are extracted and then aggregated to form the \emph{pattern-detail (PD) feature}. It promotes correlation between different regions of a point cloud and integrates the contextual information of the point cloud.  
	
	\item We propose two \emph{regulating losses} to conduct the generation of Skeypoints without any location annotations and proposals. These two regulating losses are mutually reinforcing and neither of them can be omitted.
	
	\item We conduct extensive experiments to evaluate the performance of our method, which is better than or comparable with the state-of-the-art approaches. In addition, we present an ablation test to demonstrate the advantages of the SK-Net. 
\end{itemize}

\section{Related Work}
In this section, we briefly review existing works of various regional feature extraction on point cloud and end-to-end learning of 3D keypoints.

\subsection{Extraction of regional feature on point cloud}
PointNet++\cite{pointnet++} is the earliest proposed method that addresses the problem of local information extraction. In practice, this method partitions the set of points into overlapping local regions corresponding to the spatial locations chosen by the FPS algorithm through the distance metric of the underlying space. It enables the learning of local features of a local region with increasing contextual scales. Besides, \cite{jiang2018pointsift,liu2018point2sequence} also perform the similar downsampling of a point cloud to obtain the spatial locations regarding local feature extraction. PointSIFT\cite{jiang2018pointsift} extends the traditional feature SIFT\cite{lowe2004distinctive} to develop a local region sampling approach for capturing the local information of different orientations around a spatial location. It is similar to \cite{xie2018attentional} expanding from ShapeContext\cite{belongie2001shape}.  Point2Sequence\cite{liu2018point2sequence} explores the attention-based aggregation of multi-scale areas of each local region to achieve local feature extraction. And that, \cite{hua2018pointwise} defines a pointwise convolution operator to query each point's neighbors and bin them into kernel cells for extracting pointwise local features. \cite{shen2018mining} proposes a point-set kernel operator as a set of learnable 3D points that jointly respond to a set of neighboring data points according to their geometric affinities. Other techniques also exploit extension operators to extract local structure. For example,  \cite{klokov2017escape,atzmon2018point} map the point cloud to different representation space, and \cite{simonovsky2017dynamic,wu2018dgcnn} dynamically construct a graph comprising edge features for each specific local sampling of a point cloud. In particular, \cite{le2018pointgrid,pointcnn} combine  the regular grids of 3D space with points and leverage the spatially-local correlation of regular grids to achieve local information capture. Among the latest methods, the study of local geometric relations plays an important role. \cite{lan2019modeling} models geometric local structure through decomposing the edge features along three orthogonal directions. \cite{liu2019relation} learns the geometric topology constraint among points to acquire an inductive local representation. By contrast, our method not only performs the relevant regional feature extraction in the PDE module, but also learns a set of Skeypoints corresponding to geometrically and semantically meaningful regions of the point cloud. In this way, our network can effectively model the spatial pattern of a point cloud and promotes the correlation between different regions of the point cloud. 

Most related to our approach is the recent work, SO-Net\cite{so_net}. This work suggests  utilizing the Self-Organizing Map (SOM) to build the spatial pattern of a point cloud. However, SO-Net is not proved to be generic to large-scale semantic segmentation. This is intuitively because its network is not end-to-end and the SOM is as an early stage with respect to the overall pipeline, so the SOM does not establish a connection with a specific point cloud task. In contrast, our method is capable of extending to large-scale point cloud analysis and achieving the end-to-end joint learning of Skeypoints and feature representation towards a specific point cloud task.

\subsection{End-to-end learning of 3D keypoints}
Approaches for end-to-end learning 3D keypoints have been investigated. Most of them focus on handling 3D matching or 3D pose estimation tasks. They do not extend to point cloud recognition tasks, especially classification and segmentation\cite{georgakis2018end,keypointnet,zhou2018unsupervised,endtoend2,Feng20192D3D,Lu2019DeepICP}. Some of these methods usually require keypoint proposals. For example, \cite{georgakis2018end} proposes that using a Region Proposal Network (RPN) to obtain several Region of Interests (ROIs) then determines the keypoint locations by the centroids of those ROIs. Similarly, \cite{endtoend2} casts a keypoint proposal network (KPN) comprising two convolutional layers to acquire the keypoint confidence score in evaluating whether the candidate location is a keypoint or not. In addition, KeypointNet\cite{keypointnet} presents an end-to-end geometric reasoning framework to learn an ordered set of 3D keypoints, whose discovery is guided by the carefully constructed consistency and relative pose objective functions.
These approaches tend to depend on the location consistency of keypoints of different views\cite{georgakis2018end,keypointnet,zhou2018unsupervised,endtoend2}. 
Moreover, \cite{Feng20192D3D} presents an end-to-end deep network architecture to jointly learn the descriptors for 2D and 3D keypoints from an image and point cloud, establishing 2D-3D correspondence. \cite{Lu2019DeepICP} end-to-end trains its keypoint detector to enable the system to avoid the inference of dynamic objects and leverage the help of sufficiently salient features on stationary objects. By contrast, our SK-Net focuses on point cloud classification and segmentation. In addition, our Skeypoints not only are inferred without any location annotations or proposals but also are insensitive to location consistency. This greatly benefits the spatial region relation exploration of a point cloud.

\section{End-to-end optimization of Skeypoints and feature representation}
In this section, we present the end-to-end framework of our SK-Net. First, the architecture of SK-Net is introduced. Then its important process, PDE module, is discussed in detail. Finally, the significant properties of Skeypoints are analyzed.

\subsection{SK-Net architecture}
\label{sk_archi}
\begin{figure}[t]
	\centering
	\includegraphics[width=1\columnwidth]{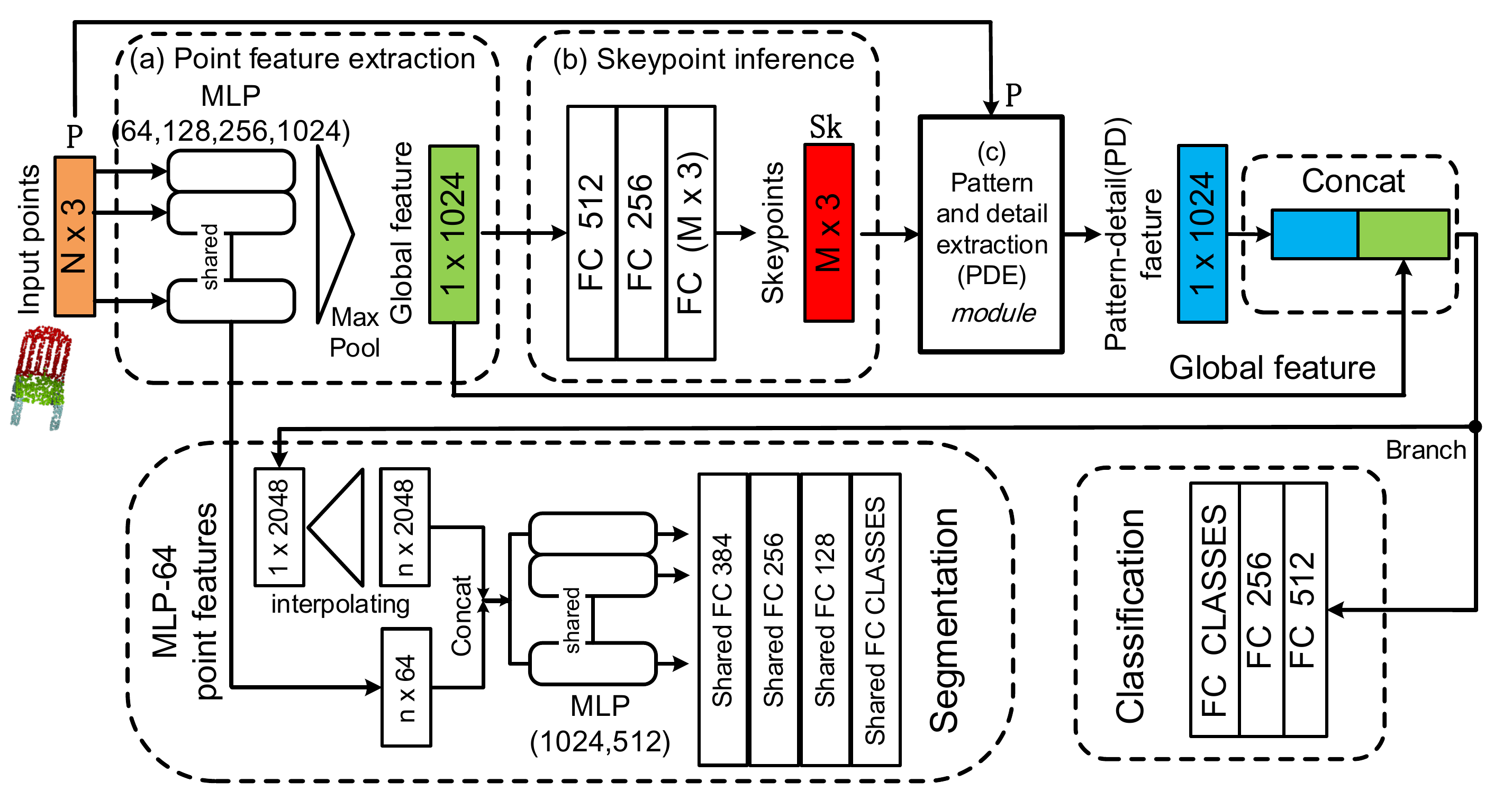} % Reduce the figure size so that it is slightly narrower than the column.
	\caption{The overall architecture of the SK-Net.}
	\label{SK_net}
\end{figure}

The overall architecture of SK-Net is shown in Figure \ref{SK_net}. The backbone of our network consists of the following three components: (a) point feature extraction, (b) skeypoint inference and (c) pattern and detail extraction. \\
(a) Point feature extraction. Given an input point cloud $P$ with size of $N$, i.e.  $P=\{p_{i}\subseteq{\mathbb{R}}^{3},i=0,1,\ldots,N-1\}$, where each point $p_{i}$ is a vector of its $(x,y,z)$ coordinates. $P$ goes through a series of multi-layer perceptrons (MLPs) to extract its point features. Next, we use the symmetry function of max-pooling to acquire the global feature. \\
(b) Skeypoint inference.  $M$ Skeypoints, $Sk=\{sk_{j}\subseteq{\mathbb{R}}^{3},j=0,1,\ldots,M-1\}$, are regressed by stacking three fully connected layers on the top of the global feature. $sk_{j}$ represents the j-th Skeypoint. Note that the MLP-64 point features are used to handle point cloud segmentation tasks. Meanwhile, the interpolating operation proposed by \cite{pointnet++} is utilized. \\
(c) Pattern and detail extraction. The generated Skeypoints are forwarded into the PDE module to get PD features. More details are presented in the next subsection. \\
Finally, our network combines the PD feature aggregated in the PDE module with the global feature obtained in (a), for a specific point cloud task e.g., classification and segmentation.

\subsection{Pattern and detail extraction module}\label{pde}
\begin{figure}[t]
	\centering
	\includegraphics[width=1\columnwidth]{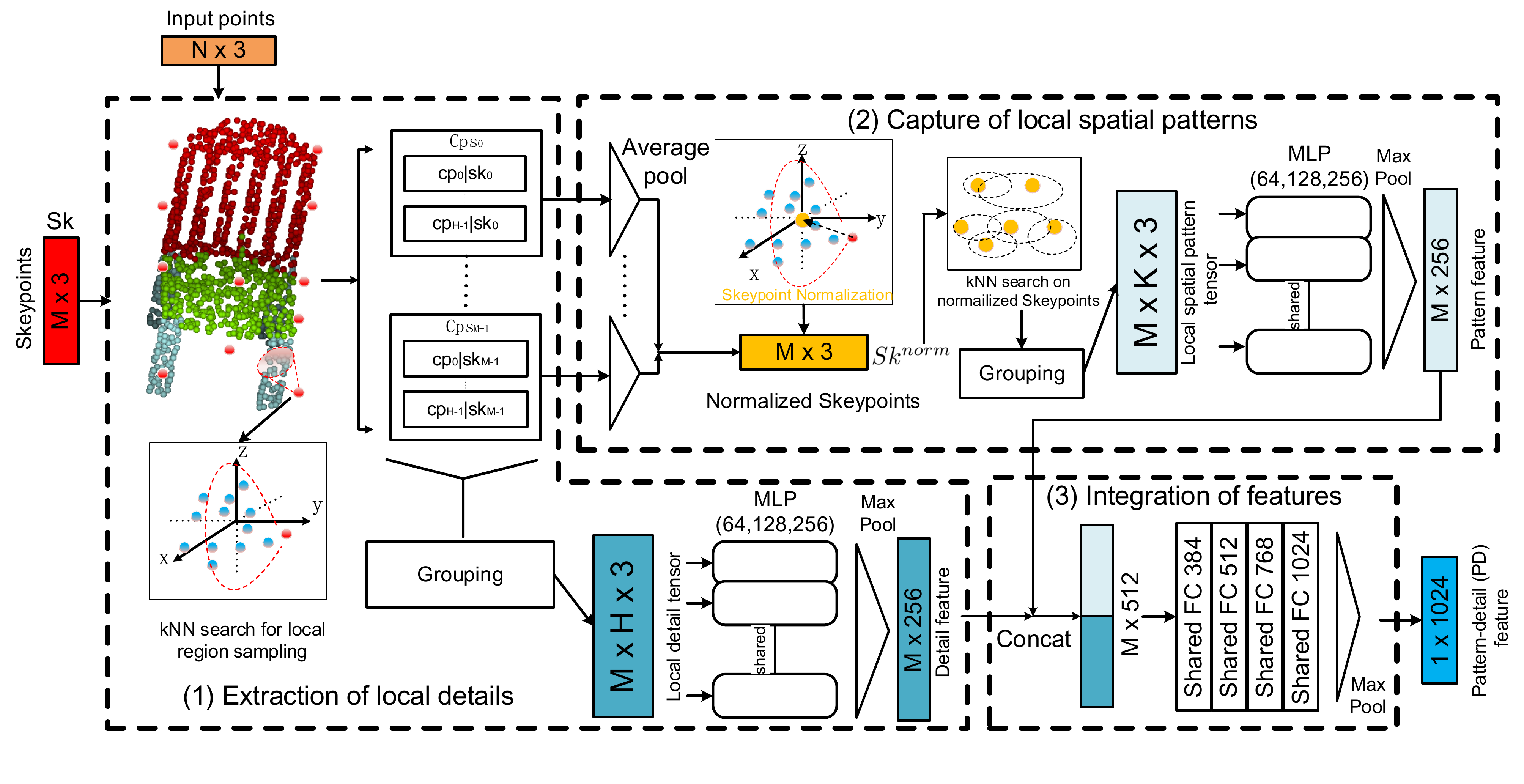}
	\caption{The pivotal PDE module of our SK-Net. The red balls around the point cloud object in (1) are the Skeypoints, and the orange balls in (2) represent the normalized Skeypoints.}\label{PDmodule}
\end{figure}

The PDE module is shown in Figure \ref{PDmodule}. The PDE module consumes $M$ generated Skeypoints and $N$ input points. It contains three parts: (1) extraction of local details, (2) capture of local spatial patterns, (3) integration of features.

\begin{figure}[htpb]
	\centering
	\includegraphics[width=1\columnwidth]{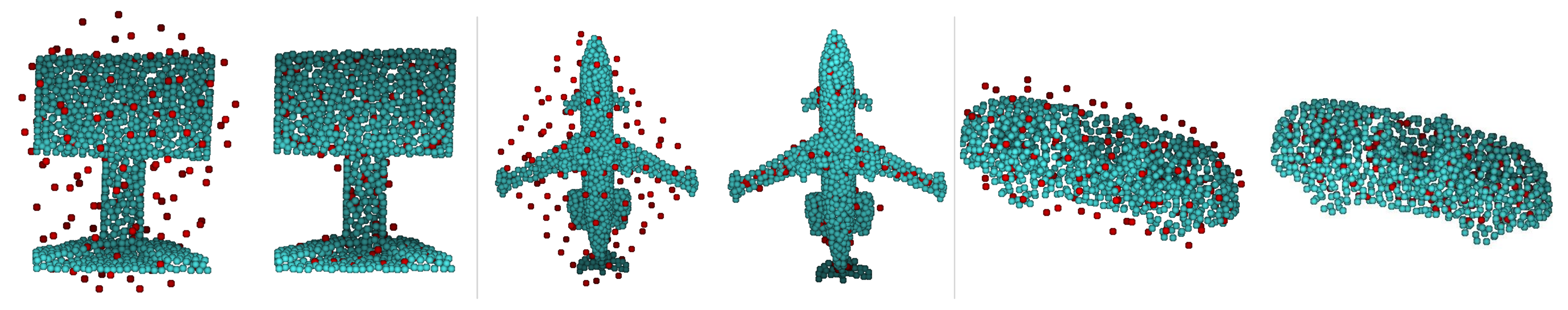} 
	\caption{Each pair contains a point cloud (the left) generated by Skeypoints and corresponding one generated by normalized Skeypoints.}
	\label{norm_origin}
\end{figure}

\paragraph{Extraction of local details}
\label{local_details}
In this part, we apply a simple kNN search to sample the local region. The number of local regions to extract local information from is identical with the number of generated Skeypoints. Consequently, $M$ local neighborhoods captured by the Skeypoints are grouped to form a $M\times{H\times{3}}$ tensor named local detail tensor. We use the $(x, y, z)$ coordinate as our tensor’s channels. The captured points are the points live in the local neighborhood captured by a generated Skeypoint and are denoted by $Cps_{j}, j=0,1,\ldots,M-1$.
$H$ is the number of captured points of each generated Skeypoint. Each captured point of a local neighborhood is denoted by $cp_{h}, h=0,1,\ldots,H-1$. The $M$ set of captured points can be formulated by:
\begin{equation}
Cps_{j} = kNN(cp_{h}|sk_{j},h=0,1,\ldots,H-1)
\end{equation}
Subsequently, PDE module applies a series of MLPs on the local detail tensor to extract the $M\times{256}$ detail feature.
It is worth mentioning that there are other ways to sample local regions. We will present relevant experiments in the next section to illustrate the superiority of using the kNN search in our scheme.
\paragraph{Capture of local spatial patterns}
To effectively capture the local spatial patterns of a point cloud, we define a concept named normalized Skeypoint. We average the location coordinates of points which live in the local neighborhood captured by each generated Skeypoint for acquiring the $M$ normalized Skeypoints, $Sk^{norm}=\{sk_{j}^{norm}\subseteq{\mathbb{R}}^{3},j=0,1,\ldots,M-1\}$, where $sk_{j}^{norm}$ is defined by:
\begin{equation}
sk_{j}^{norm} = average(Cps_{j})
\end{equation}
The normalized Skeypoints are distributed into the discriminative regions of a point cloud surface, as illustrated in Figure \ref{norm_origin}. It is beneficial to perform the entire spatial modeling of the point cloud. Next, we use kNN search on normalized Skeypoints themselves to capture the local spatial patterns of the point cloud. After that, we group the search results to get the $M\times{K\times{3}}$ local spatial pattern tensor, where $K$ is the number of normalized Skeypoints  in each local spatial region. The $M\times{256}$ pattern feature is extracted by the same approach as obtaining the detail feature.

\paragraph{Integration of features}
In this part, the PDE module concatenates the pattern feature with the detail feature to form  $M\times{512}$ intermediate result. This result is further aggregated through a stack of shared FC layers to acquire the pattern-detail (PD) feature. The PD feature builds the connections between different regions of a point cloud and integrates the contextual information of the point cloud.

\subsection{Properties of Skeypoints}
The fact that a point set is permutation invariant requires certain symmetrizations in the net computation. It is the same in our Skeypoint inference and relevant feature formation in order to perform the permutation invariance. Moreover, for our scheme and point cloud tasks, it requires that our Skeypoints are distinct from each other (distinctiveness), adapted to the spatial modeling of a point cloud (adaptation) and close to geometrically and semantically meaningful regions (closeness). However, due to the lack of keypoint annotations, the Skeypoints might either tend to be extremely close and even collapse to the same 3D location, or may stray away from the point cloud object. Because of these issues, we propose two regulating losses to conduct the generation of Skeypoints. They guarantee the distinctiveness, adaptation, and closeness of Skeypoints.

\paragraph{Permutation invariance}
We employ a modified PointNet framework to acquire the global feature of a point cloud. It is basis of the inferring of Skeypoints. After that, the activated outputs of the Skeypoint inference component are directly reshaped to obtain Skeypoints. Each coordinate of a Skeypoint is regressed independently. These procedures make Skeypoint generation satisfy permutation invariance. Note that the nonlinear activation function PReLU\cite{prelu} is adopted to guarantee that each layer activation matches the location distribution of the point cloud and the regressive consequence is well-adapted and robust. Moreover, all local features use max-pooling to ensure their permutation invariance before producing the PD feature.

\paragraph{Distinctiveness and adaptation}
In order to achieve these characteristics, we take the separation distance between Skeypoints as a hyperparameter  $\delta$ and propose the separation loss  $L_{sep}$. Hyperparameter $\delta$ provides a prior knowledge of the location distribution of the point cloud. It makes the generated Skeypoints adapted to the spatial modeling of a point cloud and boosts the convergence of our network. The separation loss $L_{sep}$ prompts the $M$ Skeypoints to be distinct from each other and penalizes two Skeypoints if their distance is less than  $\delta$. The $L_{sep}$ is formulated by:
\begin{equation}
L_{sep} = \frac{1}{M^2}\sum_{i=1}^{M}\sum_{i\neq{j}}^{M}max(0,\delta-||sk_{i}-sk_{j}||^{2})
\end{equation}

\paragraph{Closeness}
In order to make Skeypoints correspond to geometrically and semantically meaningful regions of a point cloud, the network should encourage Skeypoints to close to the discriminative regions of the point cloud surface. A regulating loss named close loss $L_{close}$ is proposed to penalize a Skeypoint and its captured points live in the local structure of this Skeypoint if they are farther than the value of hyperparameter $\theta$ in 3D Euclidean space. The $L_{close}$ is represented by: 

\begin{equation}
L_{close} = \frac{1}{MH}\sum_{i=1}^{M}\sum_{h=1}^{H}max(0,||sk_{i}-cp_{h}||^{2}-\theta)
\end{equation}

These two loss terms allow our network to discover satisfactory Skeypoints as far as possible. Besides, they prompt our network to extract more compact local features with respect to the regions corresponding to the learned Skeypoints and to capture the effective spatial pattern of a point cloud by utilizing the normalized Skeypoints.

\section{Experiments}

\paragraph{Datasets} We validate on four datasets
to demonstrate the effectiveness of our SK-Net. Object classification on ModelNet\cite{modelnet} is evaluated by accuracy, part segmentation on ShapeNetPart\cite{shapenetpart} is evaluated by mean Intersection over Union (mIoU) on points and semantic scene labeling on ScanNet\cite{scannet} is evaluated by per-point accuracy.
\begin{itemize}
	\item ModelNet. This includes two datasets which respectively contain CAD models of 10 and 40 categories. ModelNet10 consists of $4,899$ object instances which are split into $2,468$ training samples and $909$ testing samples. ModelNet40 consists of $12,311$ object instances among which $9,843$ objects belong to the training set and the other $3,991$ samples for testing. 
	\item ShapeNetPart. $16,881$ shapes from $16$ categories, labeled with $50$ parts in total. A large proportion of shape categories are annotated with two to five parts.
	\item ScanNet. $1,513$ scanned and reconstructed indoor scenes. We follow the experiment setting in \cite{pointnet++} and use 1201 scenes for training, 312 scenes for test.
\end{itemize}

\paragraph{Implementation details}
Sk-Net is implemented by Tensorflow in CUDA. We run our model on GeForce GTX Titan X for training. In general, we set the number of the Skeypoints to $192$, and $K$ to $16$, $H$ to $32$ in the PDE module. In most experiments, the hyperparameters $\delta$ , $\theta$ of our two regulating losses are both $0.05$, and the weights of all loss terms are identical. In addition, we use Adam\cite{kingma2014adam} optimizer with an initial learning rate of $0.001$ and the learning rate is decreased by staircase exponential decay. Batch size is $16$. All layers are implemented with batch normalization. PReLU activation is applied to the layers of the point feature extraction and Skeypoint inference components, while ReLU activation is applied to every layer of the subsequent network.

\subsection{Classification on ModelNet}
\label{classification}
For a fair comparison, the ModelNet10/40 datasets for our experiments are preprocessed by \cite{pointnet++}. By default, $1024$ input points are used. Moreover, we attempt to use more points and surface normals as additional features for improving performance.
Table \ref{object_classification} shows the classification accuracy of state-of-the-art methods on point cloud representation. In ModelNet10, our network outperforms state-of-the-art methods by $95.8\%$ with $1024$ points and achieves a better accuracy of $96.2\%$ with $5000$ points and normal information. In ModelNet40, the network performs a comparable result of $92.7\%$ through training with $5000$ points and surface normal vectors as additional features. Although SO-Net presents the best result in ModelNet40, its network is trained separately and not generic to large-scale semantic segmentation, while our proposed model is trained end-to-end and does extend to large-scale point cloud analysis. 

\begin{table*}[t]
	\centering
	\caption{Object classification accuracy (\%) on ModelNet.}
	\resizebox{0.6\textwidth}{!}{ % If your table exceeds the column or page width, use this command to reduce it slightly
	\begin{tabular}{llllllllll}
	\toprule
	\multicolumn{2}{l}{Method}
	& \multicolumn{2}{c}{Representation} 
	& \multicolumn{2}{c}{Input} 
	& \multicolumn{2}{c}{ModelNet10} 
	& \multicolumn{2}{c}{ModelNet40}\\
	&&&&&&Class  & Instance & Class & Instance\\ 
	\midrule
	\multicolumn{2}{l}{PointNet\cite{pointnet}}
	&\multicolumn{2}{c}{points}
	&\multicolumn{2}{c}{$1024\times{3}$}
	&\multicolumn{1}{c}{$-$}
	&\multicolumn{1}{c}{$-$}
	&\multicolumn{1}{c}{$86.2$}
	&\multicolumn{1}{c}{$89.2$}\\
	\multicolumn{2}{l}{PointNet++\cite{pointnet++}}
	&\multicolumn{2}{c}{points+normal}
	&\multicolumn{2}{c}{$5000\times{6}$}
	&\multicolumn{1}{c}{$-$}
	&\multicolumn{1}{c}{$-$}
	&\multicolumn{1}{c}{$-$}
	&\multicolumn{1}{c}{$91.9$}\\
	
	\multicolumn{2}{l}{Kd-Net\cite{klokov2017escape}}
	&\multicolumn{2}{c}{points}
	&\multicolumn{2}{c}{$2^{15}\times{3}$}
	&\multicolumn{1}{c}{$93.5$}
	&\multicolumn{1}{c}{$94.0$}
	&\multicolumn{1}{c}{$88.5$}
	&\multicolumn{1}{c}{$91.8$}\\
	
	\multicolumn{2}{l}{OctNet\cite{riegler2017octnet}}
	&\multicolumn{2}{c}{points}
	&\multicolumn{2}{c}{$128^{3}$}
	&\multicolumn{1}{c}{$90.1$}
	&\multicolumn{1}{c}{$90.9$}
	&\multicolumn{1}{c}{$83.8$}
	&\multicolumn{1}{c}{$86.5$}\\
	\multicolumn{2}{l}{SCN\cite{xie2018attentional}}
	&\multicolumn{2}{c}{points}
	&\multicolumn{2}{c}{$1024\times{3}$}
	&\multicolumn{1}{c}{$-$}
	&\multicolumn{1}{c}{$-$}
	&\multicolumn{1}{c}{$87.6$}
	&\multicolumn{1}{c}{$90.0$}\\
	\multicolumn{2}{l}{ECC\cite{simonovsky2017dynamic}}
	&\multicolumn{2}{c}{points}
	&\multicolumn{2}{c}{$1000\times{3}$}
	&\multicolumn{1}{c}{$90.0$}
	&\multicolumn{1}{c}{$90.8$}
	&\multicolumn{1}{c}{$83.2$}
	&\multicolumn{1}{c}{$87.4$}\\
	\multicolumn{2}{l}{KC-Net\cite{klokov2017escape}}
	&\multicolumn{2}{c}{points}
	&\multicolumn{2}{c}{$2048\times{3}$}
	&\multicolumn{1}{c}{$-$}
	&\multicolumn{1}{c}{$94.4$}
	&\multicolumn{1}{c}{$-$}
	&\multicolumn{1}{c}{$91.0$}\\
	\multicolumn{2}{l}{DGCNN\cite{wu2018dgcnn}}
	&\multicolumn{2}{c}{points}
	&\multicolumn{2}{c}{$1024\times{3}$}
	&\multicolumn{1}{c}{$-$}
	&\multicolumn{1}{c}{$-$}
	&\multicolumn{1}{c}{$90.2$}
	&\multicolumn{1}{c}{$92.2$}\\
	\multicolumn{2}{l}{PointGrid\cite{le2018pointgrid}}
	&\multicolumn{2}{c}{points}
	&\multicolumn{2}{c}{$1024\times{3}$}
	&\multicolumn{1}{c}{$-$}
	&\multicolumn{1}{c}{$-$}
	&\multicolumn{1}{c}{$88.9$}
	&\multicolumn{1}{c}{$92.0$}\\
	\multicolumn{2}{l}{PointCNN\cite{pointcnn}}
	&\multicolumn{2}{c}{points}
	&\multicolumn{2}{c}{$1024\times{3}$}
	&\multicolumn{1}{c}{$-$}
	&\multicolumn{1}{c}{$-$}
	&\multicolumn{1}{c}{$-$}
	&\multicolumn{1}{c}{$91.7$}\\
	\multicolumn{2}{l}{PCNN\cite{atzmon2018point}}
	&\multicolumn{2}{c}{points}
	&\multicolumn{2}{c}{$1024\times{3}$}
	&\multicolumn{1}{c}{$-$}
	&\multicolumn{1}{c}{$94.9$}
	&\multicolumn{1}{c}{$-$}
	&\multicolumn{1}{c}{$92.3$}\\
	
	\multicolumn{2}{l}{Point2Sequence\cite{liu2018point2sequence}}
	&\multicolumn{2}{c}{points}
	&\multicolumn{2}{c}{$1024\times{3}$}
	&\multicolumn{1}{c}{$95.1$}
	&\multicolumn{1}{c}{$95.3$}
	&\multicolumn{1}{c}{$90.4$}
	&\multicolumn{1}{c}{$92.6$}\\
	\multicolumn{2}{l}{SO-Net\cite{so_net}}
	&\multicolumn{2}{c}{points+normal}
	&\multicolumn{2}{c}{$5000\times{6}$}
	&\multicolumn{1}{c}{$95.5$}
	&\multicolumn{1}{c}{$95.7$}
	&\multicolumn{1}{c}{$\textbf{90.8}$}
	&\multicolumn{1}{c}{$\textbf{93.4}$}\\
	\midrule
	\multicolumn{2}{l}{Ours}
	&\multicolumn{2}{c}{points}
	&\multicolumn{2}{c}{$1024\times{3}$}
	&\multicolumn{1}{c}{$95.6$}
	&\multicolumn{1}{c}{$95.8$}
	&\multicolumn{1}{c}{$89.0$}
	&\multicolumn{1}{c}{$91.5$}\\
	
	\multicolumn{2}{l}{Ours}
	&\multicolumn{2}{c}{points+normal}
	&\multicolumn{2}{c}{$5000\times{6}$}
	&\multicolumn{1}{c}{$\textbf{96.2}$}
	&\multicolumn{1}{c}{$\textbf{96.2}$}
	&\multicolumn{1}{c}{$90.3$}
	&\multicolumn{1}{c}{$92.7$}\\
	\bottomrule
	\\
	\end{tabular}}
	%}
	\label{object_classification}
\end{table*}

\subsection{Part segmentation on ShapeNetPart}
\label{part_seg}
Part segmentation is more challenging than object classification and can be formulated as a per-point classification problem. We use the segmentation network to do this. Fairly, ShapeNetPart dataset is prepared by \cite{pointnet++} and we also follow the metric protocol from \cite{pointnet++}.
The results of the related approaches are illustrated in Table \ref{object_part_seg}. Although the best mIoU of all shapes is performed by \cite{liu2018point2sequence}, ours outperforms state-of-the-art methods in six categories and achieves comparable results in the remaining categories. We provide ShapeNetPart segmentation visualization in the supplementary material.

\begin{table*}[htpb]
	\centering
	\caption{Object part segmentation results on ShapeNetPart dataset.}
	\resizebox{0.75\textwidth}{!}{ 
\begin{tabular}{llllllllllllllllllll}
	\toprule
	\multicolumn{3}{c}{}
	&\multicolumn{17}{c}{\small Intersection over Union (IoU)}\\
	\cmidrule(l){4-20}
	\multicolumn{3}{l}{}
	&\multicolumn{1}{c}{\textbf{mean}}
	&\multicolumn{1}{c}{air}
	&\multicolumn{1}{c}{bag}
	&\multicolumn{1}{c}{cap}
	&\multicolumn{1}{c}{car}
	&\multicolumn{1}{c}{chair}
	&\multicolumn{1}{c}{ear.}
	&\multicolumn{1}{c}{gui.}
	&\multicolumn{1}{c}{knife}
	&\multicolumn{1}{c}{lamp}
	&\multicolumn{1}{c}{lap.}
	&\multicolumn{1}{c}{motor}
	&\multicolumn{1}{c}{mug}
	&\multicolumn{1}{c}{pistol}
	&\multicolumn{1}{c}{rocket}
	&\multicolumn{1}{c}{skate}
	&\multicolumn{1}{c}{table}
	\\ 
	\midrule
	\multicolumn{3}{l}{PointNet\cite{pointnet}}
	&\multicolumn{1}{c}{$83.7$}
	&\multicolumn{1}{c}{$83.4$}
	&\multicolumn{1}{c}{$78.7$}
	&\multicolumn{1}{c}{$82.5$}
	&\multicolumn{1}{c}{$74.9$}
	&\multicolumn{1}{c}{$89.6$}
	&\multicolumn{1}{c}{$73.0$}
	&\multicolumn{1}{c}{$\textbf{91.5}$}
	&\multicolumn{1}{c}{$85.9$}
	&\multicolumn{1}{c}{$80.8$}
	&\multicolumn{1}{c}{$95.3$}
	&\multicolumn{1}{c}{$65.2$}
	&\multicolumn{1}{c}{$93.0$}
	&\multicolumn{1}{c}{$81.2$}
	&\multicolumn{1}{c}{$57.9$}
	&\multicolumn{1}{c}{$72.8$}
	&\multicolumn{1}{c}{$80.6$}
	\\
	\multicolumn{3}{l}{PointNet++\cite{pointnet++}}
	
	&\multicolumn{1}{c}{$85.1$}
	&\multicolumn{1}{c}{$82.4$}
	&\multicolumn{1}{c}{$79.0$}
	&\multicolumn{1}{c}{$87.7$}
	&\multicolumn{1}{c}{$77.3$}
	&\multicolumn{1}{c}{$90.8$}
	&\multicolumn{1}{c}{$71.8$}
	&\multicolumn{1}{c}{$91.0$}
	&\multicolumn{1}{c}{$85.9$}
	&\multicolumn{1}{c}{$83.7$}
	&\multicolumn{1}{c}{$95.3$}
	&\multicolumn{1}{c}{$\textbf{71.6}$}
	&\multicolumn{1}{c}{$94.1$}
	&\multicolumn{1}{c}{$81.3$}
	&\multicolumn{1}{c}{$58.7$}
	&\multicolumn{1}{c}{$\textbf{76.4}$}
	&\multicolumn{1}{c}{$82.6$}
	\\
	\multicolumn{3}{l}{Kd-Net\cite{klokov2017escape}}
	&\multicolumn{1}{c}{$82.3$}
	&\multicolumn{1}{c}{$80.1$}
	&\multicolumn{1}{c}{$74.6$}
	&\multicolumn{1}{c}{$74.3$}
	&\multicolumn{1}{c}{$70.3$}
	&\multicolumn{1}{c}{$88.6$}
	&\multicolumn{1}{c}{$73.5$}
	&\multicolumn{1}{c}{$90.2$}
	&\multicolumn{1}{c}{$87.2$}
	&\multicolumn{1}{c}{$81.0$}
	&\multicolumn{1}{c}{$94.9$}
	&\multicolumn{1}{c}{$57.4$}
	&\multicolumn{1}{c}{$86.7$}
	&\multicolumn{1}{c}{$78.1$}
	&\multicolumn{1}{c}{$51.8$}
	&\multicolumn{1}{c}{$69.9$}
	&\multicolumn{1}{c}{$80.3$}
	\\
	\multicolumn{3}{l}{KC-Net\cite{shen2018mining}}
	&\multicolumn{1}{c}{$84.7$}
	&\multicolumn{1}{c}{$82.8$}
	&\multicolumn{1}{c}{$81.5$}
	&\multicolumn{1}{c}{$86.4$}
	&\multicolumn{1}{c}{$77.6$}
	&\multicolumn{1}{c}{$90.3$}
	&\multicolumn{1}{c}{$76.8$}
	&\multicolumn{1}{c}{$91.0$}
	&\multicolumn{1}{c}{$87.2$}
	&\multicolumn{1}{c}{$\textbf{84.5}$}
	&\multicolumn{1}{c}{$95.5$}
	&\multicolumn{1}{c}{$69.2$}
	&\multicolumn{1}{c}{$94.4$}
	&\multicolumn{1}{c}{$81.6$}
	&\multicolumn{1}{c}{$60.1$}
	&\multicolumn{1}{c}{$75.2$}
	&\multicolumn{1}{c}{$81.3$}
	\\
	\multicolumn{3}{l}{DGCNN \cite{wu2018dgcnn}}
	&\multicolumn{1}{c}{$85.1$}
	&\multicolumn{1}{c}{$\textbf{84.2}$}
	&\multicolumn{1}{c}{$\textbf{83.7}$}
	&\multicolumn{1}{c}{$84.4$}
	&\multicolumn{1}{c}{$77.1$}
	&\multicolumn{1}{c}{$\textbf{90.9}$}
	&\multicolumn{1}{c}{$78.5$}
	&\multicolumn{1}{c}{$\textbf{91.5}$}
	&\multicolumn{1}{c}{$87.3$}
	&\multicolumn{1}{c}{$82.9$}
	&\multicolumn{1}{c}{$96.0$}
	&\multicolumn{1}{c}{$67.8$}
	&\multicolumn{1}{c}{$93.3$}
	&\multicolumn{1}{c}{$\textbf{82.6}$}
	&\multicolumn{1}{c}{$59.7$}
	&\multicolumn{1}{c}{$75.5$}
	&\multicolumn{1}{c}{$82.0$}
	\\

	\multicolumn{3}{l}{point2sequence \cite{liu2018point2sequence}}
	&\multicolumn{1}{c}{$\textbf{85.2}$}
	&\multicolumn{1}{c}{$82.6$}
	&\multicolumn{1}{c}{$81.8$}
	&\multicolumn{1}{c}{$87.5$}
	&\multicolumn{1}{c}{$77.3$}
	&\multicolumn{1}{c}{$90.8$}
	&\multicolumn{1}{c}{$77.1$}
	&\multicolumn{1}{c}{$91.1$}
	&\multicolumn{1}{c}{$86.9$}
	&\multicolumn{1}{c}{$83.9$}
	&\multicolumn{1}{c}{$\textbf{95.7}$}
	&\multicolumn{1}{c}{$70.8$}
	&\multicolumn{1}{c}{$\textbf{94.6}$}
	&\multicolumn{1}{c}{$79.3$}
	&\multicolumn{1}{c}{$58.1$}
	&\multicolumn{1}{c}{$75.2$}
	&\multicolumn{1}{c}{$82.8$}
	\\
	\multicolumn{3}{l}{SO-Net \cite{so_net}}
	&\multicolumn{1}{c}{$84.9$}
	&\multicolumn{1}{c}{$82.8$}
	&\multicolumn{1}{c}{$77.8$}
	&\multicolumn{1}{c}{$\textbf{88.0}$}
	&\multicolumn{1}{c}{$77.3$}
	&\multicolumn{1}{c}{$90.6$}
	&\multicolumn{1}{c}{$73.5$}
	&\multicolumn{1}{c}{$90.7$}
	&\multicolumn{1}{c}{$83.9$}
	&\multicolumn{1}{c}{$82.8$}
	&\multicolumn{1}{c}{$94.8$}
	&\multicolumn{1}{c}{$69.1$}
	&\multicolumn{1}{c}{$94.2$}
	&\multicolumn{1}{c}{$80.9$}
	&\multicolumn{1}{c}{$53.1$}
	&\multicolumn{1}{c}{$72.9$}
	&\multicolumn{1}{c}{$\textbf{83.0}$}
	\\
	\midrule
	\multicolumn{3}{l}{Ours}
	&\multicolumn{1}{c}{$85.0$}
	&\multicolumn{1}{c}{$82.9$}
	&\multicolumn{1}{c}{$80.7$}
	&\multicolumn{1}{c}{$87.6$}
	&\multicolumn{1}{c}{$\textbf{77.8}$}
	&\multicolumn{1}{c}{$90.5$}
	&\multicolumn{1}{c}{$\textbf{79.9}$}
	&\multicolumn{1}{c}{$91.0$}
	&\multicolumn{1}{c}{$\textbf{88.1}$}
	&\multicolumn{1}{c}{$84.0$}
	&\multicolumn{1}{c}{$\textbf{95.7}$}
	&\multicolumn{1}{c}{$69.9$}
	&\multicolumn{1}{c}{$94.0$}
	&\multicolumn{1}{c}{$81.1$}
	&\multicolumn{1}{c}{$\textbf{60.8}$}
	&\multicolumn{1}{c}{$\textbf{76.4}$}
	&\multicolumn{1}{c}{$81.9$}
	\\
	\bottomrule
	\\
\end{tabular}}
	%}
	\label{object_part_seg}
\end{table*}

\subsection{Semantic scene labeling on ScanNet}
We conduct experiments on ScanNet’s Semantic Scene Labeling task to validate that our method is suitable for large-scale point cloud analysis. We use our segmentation network to do this. To perform this task, we set the point cloud to be normalized into $[-1,1]$ because the $L_{sep}$ is sensitive to the distribution of point clouds. All experiments use $8192$ points. We compare the per-point accuracy and the number of the local region features in each method, as shown in Table \ref{scan}. Our approach is better than PointNet and slightly behind PointNet++. However, the number of our local region features is only $192$ which is much smaller than that in PointNet++. These results show that our method is extendable to large-scale semantic segmentation and the local region features obtained by our method are compact and efficient.

\subsection{Ablation study}
In this subsection, we further conduct several ablation experiments to investigate various setup variations and to demonstrate the advantages of SK-Net.  

\paragraph{Effects of features extracted in PDE module}
We present an ablation test on ModelNet10 classification to show the effect of the features extracted by PDE module i.e. detail features and pattern features, as shown in Figure \ref{PDmodule}. The accuracy of the experiments as follows: $93.9\%$ (using only detail features), $93.3\%$ (using only pattern features), and $95.8\%$ (using their concatenated result). It demonstrates that our proposed PD feature promotes correlation between different regions of a point cloud and is more effective in modeling the whole spatial distribution of the point cloud.
\begin{table}[htpb]
	\caption{Comparisons of per-point classification on ScanNet.} \label{scan}
	\centering
	\resizebox{0.9\columnwidth}{!}{\begin{tabular}{llll}
			\toprule
			\multicolumn{1}{c}{Method}
			& \multicolumn{1}{|c}{Local regions}
			& \multicolumn{1}{c|}{Local selection operator}  
			& \multicolumn{1}{c}{accuracy} 
			\\
			\midrule
			\multicolumn{1}{c}{PointNet}
			&\multicolumn{1}{|c}{$-$}
			&\multicolumn{1}{c|}{$-$}
			&\multicolumn{1}{c}{$77.7\%$}\\
			
			\multicolumn{1}{c}{PointNet++(ssg)}
			&\multicolumn{1}{|c}{$[1024,256,64,16]$}
			&\multicolumn{1}{c|}{FPS}
			&\multicolumn{1}{c}{$82.6\%$}\\
			
			\multicolumn{1}{c}{PointNet++(msg)}
			&\multicolumn{1}{|c}{$[512,128]$}
			&\multicolumn{1}{c|}{FPS}
			&\multicolumn{1}{c}{$83.2\%$}\\
			\midrule
			\multicolumn{1}{c}{Ours}
			&\multicolumn{1}{|c}{$192$}
			&\multicolumn{1}{c|}{End-to-end learning}
			&\multicolumn{1}{c}{$81.4\%$}\\
			
			\bottomrule
			\\
	\end{tabular}}
	
\end{table}

\paragraph{Complementarity of regulating losses}
Ablation experiments of the losses are presented to validate the effect of our
loss functions (classification loss: $L_{cls}$, separation loss:  $L_{sep}$, close loss:  $L_{close}$). As shown in Table \ref{losses}, EXP.1 is the baseline and only the classification loss is used. The results show that our method is effective. EXP.2,3,4 show that the two regulating losses proposed by our scheme are mutually reinforcing and neither of them can be omitted. The effect of combining these two losses can refine the spatial model and improve the performance by $1.5\%$.

\begin{table}[htpb]
	\caption{Ablation test of the losses on ModelNet10
		classification.}
	\centering
	\resizebox{.5\columnwidth}{!}{
		\begin{tabular}{lllll}
			\toprule
			\multicolumn{1}{c}{EXP.}
			& \multicolumn{1}{c}{$L_{cls}$} 
			& \multicolumn{1}{c}{$L_{sep}$} 
			& \multicolumn{1}{c}{$L_{close}$} 
			& \multicolumn{1}{c}{accuracy}\\
			\midrule
			\multicolumn{1}{c}{1}
			&\multicolumn{1}{c}{\checkmark}
			&\multicolumn{1}{c}{$-$}
			&\multicolumn{1}{c}{$-$}
			&\multicolumn{1}{c}{$94.3\%$}\\
			\midrule
			\multicolumn{1}{c}{2}
			&\multicolumn{1}{c}{\checkmark}
			&\multicolumn{1}{c}{\checkmark}
			&\multicolumn{1}{c}{$-$}
			&\multicolumn{1}{c}{$93.8\%$}\\
			\midrule
			\multicolumn{1}{c}{3}
			&\multicolumn{1}{c}{\checkmark}
			&\multicolumn{1}{c}{$-$}
			&\multicolumn{1}{c}{\checkmark}
			&\multicolumn{1}{c}{$93.2\%$}\\
			\midrule
			\multicolumn{1}{c}{4}
			&\multicolumn{1}{c}{\checkmark}
			&\multicolumn{1}{c}{\checkmark}
			&\multicolumn{1}{c}{\checkmark}
			&\multicolumn{1}{c}{$\textbf{95.8\%}$}\\
			\bottomrule
			\\
	\end{tabular}}
	\label{losses}
\end{table}

\paragraph{Effects of local region samplings}
In this part, we experiment using other techniques (ball query and sift query proposed by \cite{jiang2018pointsift}) to sample local regions and also play with the general search radii: $0.1, 0.2$. For all experiments, the number of sample points is $32$ and the number of input points is $1024$. As shown in Table \ref{neigh},  the sift query is the least effective method and the ball query is slightly worse than kNN. This shows kNN is the most beneficial to our scheme.
\begin{table}[htpb]
	\caption{Effects of local sample choices on
		ModelNet10 classification.} 
	\centering
	\resizebox{.6\columnwidth}{!}{
		\begin{tabular}{lllll}
			\toprule
			\multicolumn{1}{c}{kNN}
			& \multicolumn{2}{|c|}{ball query} 
			& \multicolumn{2}{c}{sift query}\\
			\multicolumn{1}{c}{}
			& \multicolumn{1}{|c}{r=0.1} 
			& \multicolumn{1}{c|}{r=0.2} 
			& \multicolumn{1}{c}{r=0.1} 
			& \multicolumn{1}{c}{r=0.2} \\
			\midrule
			\multicolumn{1}{c}{$\textbf{95.8\%}$}
			&\multicolumn{1}{|c}{$94.3\%$}
			&\multicolumn{1}{c|}{$94.7\%$}
			&\multicolumn{1}{c}{$92.5\%$}
			&\multicolumn{1}{c}{$92.7\%$}\\
			\bottomrule
			\\
	\end{tabular}}
	\label{neigh}
\end{table}
\paragraph{Effectiveness of Skeypoints}

\begin{figure}[htpb]
\centering
\includegraphics[width=0.75\columnwidth]{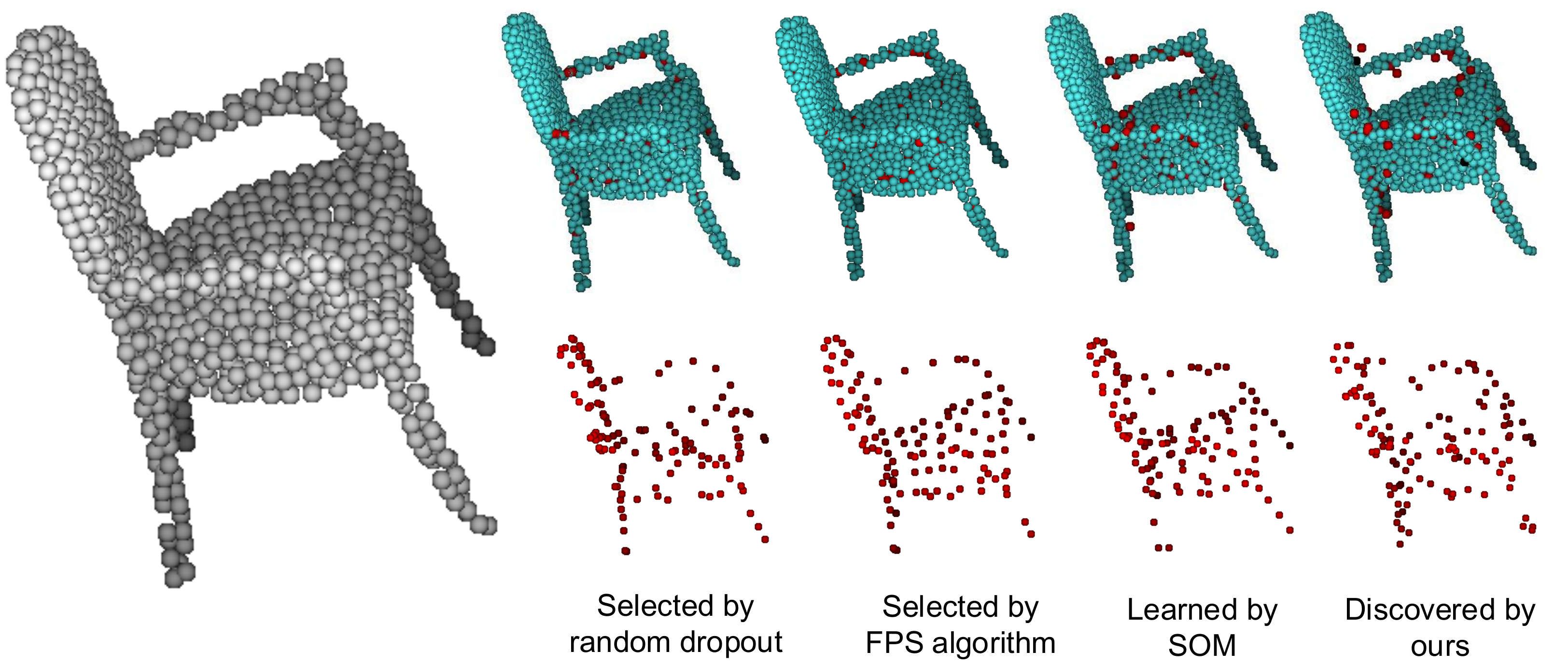} 
\caption{There is a point cloud on the left-most side and the corresponding normalized keypoints(red balls) discovered by different methods on the other side.}
	\label{effectiveness_all_visual}
\end{figure}

\begin{figure}[htpb]
	\centering	
\begin{tikzpicture}[scale=0.5]
\begin{axis}[
legend style={at={(0.6,0.2)},anchor=east},
xlabel={Number of points},
ylabel={Accuracy},
x dir=reverse
]

\addplot[ultra thick,smooth,color=blue,mark=x]
coordinates {
	(1024,91.4) (768,90.4)	(512,74.1) (256,31.5) 
};
\addlegendentry{Random dropout}
\addplot[ultra thick,smooth,color=orange,mark=x]
coordinates {
	(1024,92.2) (768,89.2) (512,60.9) (256,20.1)
};
\addlegendentry{FPS algorithm}
\addplot[ultra thick,smooth,mark=x,color=green] coordinates {
	(1024,93.8) (768,93.3) (512,90.4) (256,77.2)
};
\addlegendentry{SOM}
\addplot[ultra thick,smooth,mark=x,color=red] coordinates {
	(1024,95.5) (768,94.5) (512,92.2) (256,75.1)
};
\addlegendentry{Ours}
\end{axis}
\end{tikzpicture}
	\caption{Curve shows the performance of the spatial modeling approaches on various points densities.}	
	\label{effectiveness_all_curve}
\end{figure}
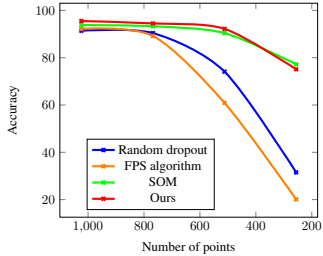

In order to exhibit the effectiveness of our Skeypoints, we eliminate the Skeypoints. And then the network is trained respectively with random dropout, farthest point sampling (FPS) algorithm and the SOM. We use these methods to discover a set of keypoint-like points for building the spatial pattern of a point cloud. Note that the keypoints are jittered slightly to avoid turning into naive downsampling in the random dropout and FPS methods. In addition, we use $11\times{11}$ SOM nodes processed by SO-Net\cite{so_net} and likewise the number of keypoints is $121$ for all comparative experiments. Subsequently, we evaluate the accuracy of different spatial modeling methods on ModelNet10. Visualization examples and the accuracy results are illustrated in Figure \ref{effectiveness_all_visual} and Figure \ref{effectiveness_all_curve}. 
It is interesting that the keypoints respectively selected by the FPS algorithm and learned by SOM, are more well-distributed than our method but have approximately $2$-$3$ percent lower accuracy than ours. Moreover, it is more sensitive to the density perturbation of points with the unlearned random dropout and FPS algorithm. By contrast, our Skeypoints are jointly optimized with the feature learning for a specific point cloud task, promoting the robustness of the variability of points density and performing better performance.

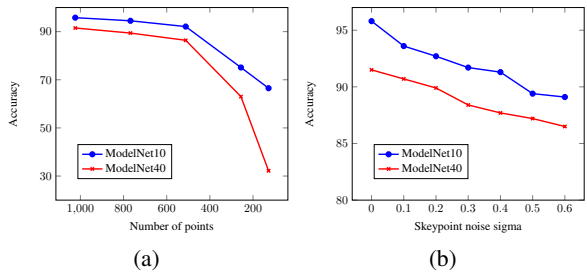
\begin{figure}[htpb]
	\centering
	\subfloat[]{
		\centering
		\begin{tikzpicture}[scale=0.45]
		\begin{axis}[
		legend style={at={(0.5,0.2)},anchor=east},
		ytick={90,70,50,30},
		xlabel={Number of points},
		ylabel={Accuracy},
		x dir = reverse,
		ymin=20,
		ymax=100
		]
		\addplot[very thick,mark=*,blue] coordinates {
			(128,66.5)(256,75.1) (512,92.1) (768,94.5) (1024,95.8)
		};
		\addlegendentry{ModelNet10}
		\addplot[very thick,color=red,mark=x]
		coordinates {
			(128,32.2) (256,63.0) (512,86.4) (768,89.4) (1024,91.5)
		};
		\addlegendentry{ModelNet40}
		\end{axis}
		\end{tikzpicture}
	}
	\subfloat[]{
		\centering
		\begin{tikzpicture}[scale=0.45]
		\begin{axis}[
		legend style={at={(0.5,0.2)},anchor=east},
		ytick={80,85,90,95},
		xlabel={Skeypoint noise sigma},
		ylabel={Accuracy},
		ymin=80,
		ymax=97
		]
		\addplot[very thick,mark=*,blue] coordinates {
			(0,95.8) (0.1,93.6) (0.2,92.7) (0.3,91.7)
			(0.4,91.3)(0.5,89.4)
			(0.6,89.1)
		};
		\addlegendentry{ModelNet10}
		\addplot[very thick,color=red,mark=x]
		coordinates {
			(0,91.5) (0.1,90.7) (0.2,89.9) (0.3,88.4)
			(0.4,87.7)(0.5,87.2)
			(0.6,86.5)
		};
		\addlegendentry{ModelNet40}
		\end{axis}
		\end{tikzpicture}
	}

	\caption{Robustness to the corruption. (a) There is point dropout randomly during testing, and the size $H$ decrease by $8$ when the number of input points is less than $512$. (b) Gaussian noise $\mathcal N (0,\sigma)$ is added to the generated Skeypoints during testing.}
	\label{robust} 
\end{figure}

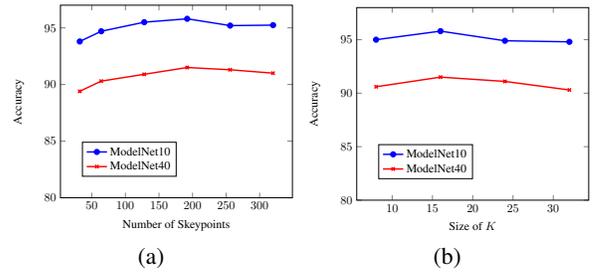
\begin{figure}[htpb]
	\centering
	\subfloat[]{
		\centering
		\begin{tikzpicture}[scale=0.45]
		\begin{axis}[
		legend style={at={(0.5,0.2)},anchor=east},
		ytick={80,85,90,95},
		xlabel={Number of Skeypoints},
		ylabel={Accuracy},
		ymin=80,
		ymax=97
		]
		\addplot[very thick,mark=*,blue] coordinates {
			(32,93.8) (64,94.7) (128,95.5) (192,95.8)(256,95.2)(320,95.24)
		};
		\addlegendentry{ModelNet10}
		\addplot[very thick,color=red,mark=x]
		coordinates {
			(32,89.4) (64,90.3) (128,90.9) (192,91.5)(256,91.3)(320,91.0)
		};
		\addlegendentry{ModelNet40}
		\end{axis}
		\end{tikzpicture}
	}
	\subfloat[]{
		\centering
		\begin{tikzpicture}[scale=0.45]
		\begin{axis}[
		legend style={at={(0.5,0.2)},anchor=east},
		ytick={80,85,90,95},
		xlabel={Size of $K$},
		ylabel={Accuracy},
		ymin=80,
		ymax=98
		]
		\addplot[very thick,mark=*,blue] coordinates {
			(8,95.0)
			(16,95.8)
			(24,94.9) (32,94.8)
			
		};
		\addlegendentry{ModelNet10}
		\addplot[very thick,color=red,mark=x]
		coordinates {
			(8,90.6) (16,91.5) (24,91.1)(32,90.3)
		};
		\addlegendentry{ModelNet40}
		\end{axis}
		\end{tikzpicture}
	}
	\caption{Effect of preferences. (a) The effect of the number of Skeypoints on the condition that $K$ is $16$. (b) The effect of various $K$ when fixing $128$ Skeypoints.}
	\label{preference} 
\end{figure}

\paragraph{Robustness to corruption}
The network is trained with input points of $1024$ and evaluated with point dropout randomly to demonstrate the robustness of the point density variability. In ModelNet10/40, our accuracy drops by 3.4\% and 4.5\% respectively with 50\% points missing (1024 to 512), and remains 75.1\% and 63.0\% respectively with $256$ points, as shown in Figure \ref{robust} (a). Moreover, our SK-Net is robust to the noise or corruption of the Skeypoints, as exhibited in Figure \ref{robust} (b). When the noise sigma is the maximum of $0.6$ in our experiments, the accuracy on ModelNet10/40 is respectively $89.1$ and $86.5$.
\paragraph{Effects of preferences}
Our proposed method is more than simply using the end-to-end learned Skeypoints to choose the local regions for extracting the local features, as an existing method\cite{pointnet++} does. The Skeypoints also play an important role in the spatial modeling of a point cloud, enhancing the regional associations and prompting our model to get more compact regional features, as illustrated in Table \ref{scan}.  Therefore, the number of generated Skeypoints should be adapted to urge the Skeypoints to embrace the point cloud adequately, and the size $K$ of the normalized Skeypoints of a local spatial region should properly reinforce the region correlation of the point cloud. As shown in Figure \ref{preference} (a), our network gradually performs better accuracy with the Skeypoints increasing from $32$ to $192$, while the accuracy decreases slightly with $256$ Skeypoints. And then the effect of the size $K$ is also shown in Figure \ref{preference} (b) on condition that the number of generated Skeypoints remains at $128$. The best accuracies are achieved when $K$ is 16.

\section{Conclusions}
In this work, we propose an end-to-end framework named SK-Net to jointly optimize the inference of Skeypoints with the feature learning of a point cloud. These Skeypoints are generated by two complementary regulating losses. Specifically, their generation requires neither location labels and proposals nor location consistency. Furthermore, we design the PDE module to extract and integrate the detail feature and pattern feature, so that local region feature extraction and the spatial modeling of a point cloud can be achieved efficiently. As a result, the correlation between different regions of a point cloud is enhanced, and the learning of the contextual information of the point cloud is promoted. In addition, we conduct experiments to show that for both point cloud classification and segmentation, our method achieves better or similar performance when compared with other existing methods. The advantages of our SK-Net is also demonstrated by several ablation experiments.

\section*{Acknowledgement}
Yunqi Lei is the corresponding author.
This work was supported by the National Natural Science Foundation of China (61671397). We thank all anonymous reviewers for their constructive comments.

\small
\bibliography{2180-references}
\bibliographystyle{aaai}
\end{document}